\documentclass[runningheads]{llncs}

\usepackage{eccv}

\usepackage{eccvabbrv}

\usepackage{graphicx}
\usepackage{booktabs}
\usepackage{colortbl}
\usepackage{inconsolata}
\usepackage{multirow}
\usepackage{algorithm}
\usepackage{subcaption}
\usepackage{amsmath}
\usepackage{float}
\usepackage{amssymb}
\usepackage{pifont}
\newcommand{\cmark}{\ding{51}}%
\newcommand{\xmark}{\ding{55}}%
\usepackage{wrapfig}
\usepackage[dvipsnames]{xcolor}
\usepackage[accsupp]{axessibility}  

\usepackage{hyperref}

\usepackage{orcidlink}
\usepackage{microtype}
\definecolor{Mycolor1}{HTML}{8FAADC}
\definecolor{cYellow}{HTML}{FFFFCC}
\definecolor{cRed}{HTML}{FFCCCC} 
\definecolor{cGrey}{HTML}{F3F7F2} 
\definecolor{cGreen}{HTML}{548235}

\hypersetup{
    colorlinks=true,       
    linkcolor=red,          
    citecolor=eccvblue,        
}

\begin{document}

\title{SQ-LLaVA: Self-Questioning for Large Vision-Language Assistant} 


\author{Guohao Sun\inst{1}\orcidlink{0009-0002-0935-6196} \and
Can Qin\inst{2}\orcidlink{0000-0003-0712-5378} \and
Jiamian Wang\inst{1}\orcidlink{0000-0002-0074-0274} \and
Zeyuan Chen\inst{2}\orcidlink{0009-0003-2471-5449} \and \\
Ran Xu\inst{2} \orcidlink{0009-0004-4585-5261}\and
Zhiqiang Tao\inst{1}\orcidlink{0000-0002-5639-7540}
}

\authorrunning{G.~Sun et al.}

\institute{Rochester Institute of Technology, Rochester, NY, US  \and
Salesforce AI Research, CA, US
}

\maketitle

\begin{abstract}
Recent advances in vision-language models have shown notable generalization in broad tasks through visual instruction tuning. However, bridging the gap between the pre-trained vision encoder and the large language models (LLMs) becomes the whole network's bottleneck. To improve cross-modality alignment, existing works usually consider more visual instruction data covering a broader range of vision tasks to fine-tune the model for question-answering, which, however, is costly to obtain and has not thoroughly explored the rich contextual information contained in images. This paper first attempts to harness the overlooked context within visual instruction data, training the model to self-supervised ``learning'' how to ask high-quality questions. In this way, we introduce a novel framework named SQ-LLaVA: Self-Questioning for Large Vision-Language Assistant. SQ-LLaVA exhibits proficiency in generating flexible and meaningful image-related questions while analyzing the visual clue and prior language knowledge, signifying an advanced level of generalized visual understanding. Moreover, fine-tuning SQ-LLaVA on higher-quality instruction data shows a performance improvement compared with traditional visual-instruction tuning methods. This improvement highlights the efficacy of self-questioning techniques in achieving a deeper and more nuanced comprehension of visual content across various contexts. Our code is available at \url{https://github.com/heliossun/SQ-LLaVA}. 
\end{abstract}


\section{Introduction}
\label{sec:intro}
The recently emerging large vision-language methods, such as large language-and-vision assistant (LLaVA) and its variants~\cite{llava,llava1-5, Chen2023ShareGPT4VIL,Bai2023QwenVLAV,zhang2023llama}, fine-tune large language models (LLM)~\cite{vicuna, zhang2023llama, Bai2023QwenVLAV} on visual instruction data~\cite{Zhao2023SVITSU, zhu2023minigpt, Zhang2023LLaVAREV,li2023empowering} to realize diverse open-world multimodal understanding, demonstrating a surprising efficacy of \emph{visual instruction tuning} -- the LLM learns to perform complex vision tasks by conditioning on a prompt containing image and text clues. Existing visual instruction data is mainly built upon conversations (e.g., ChatGPT/GPT4-V~\cite{2023GPT4VisionSC}), consisting of images and multiple question-answer pairs. Building high-quality visual instruction data usually requires images and texts from different tasks to generate diverse questions, such as ``Please provide the bounding box coordinate of the region this sentence describes A dead leaf on the ground'' for the object detection task. Empirically, by increasing the diversity of questions, LLaVA has achieved better performance on GQA and VizWiz tasks (26\% and 45\% over previous state-of-the-art methods~\cite{Dai2023InstructBLIPTG}). This evidence strongly suggests the advantage of training models on a broad spectrum and diverse array of tasks for enriching general vision-language understanding.

\begin{figure}[t]
\centering
\begin{subfigure}{.62\textwidth}
  \centering
  \includegraphics[width=\linewidth]{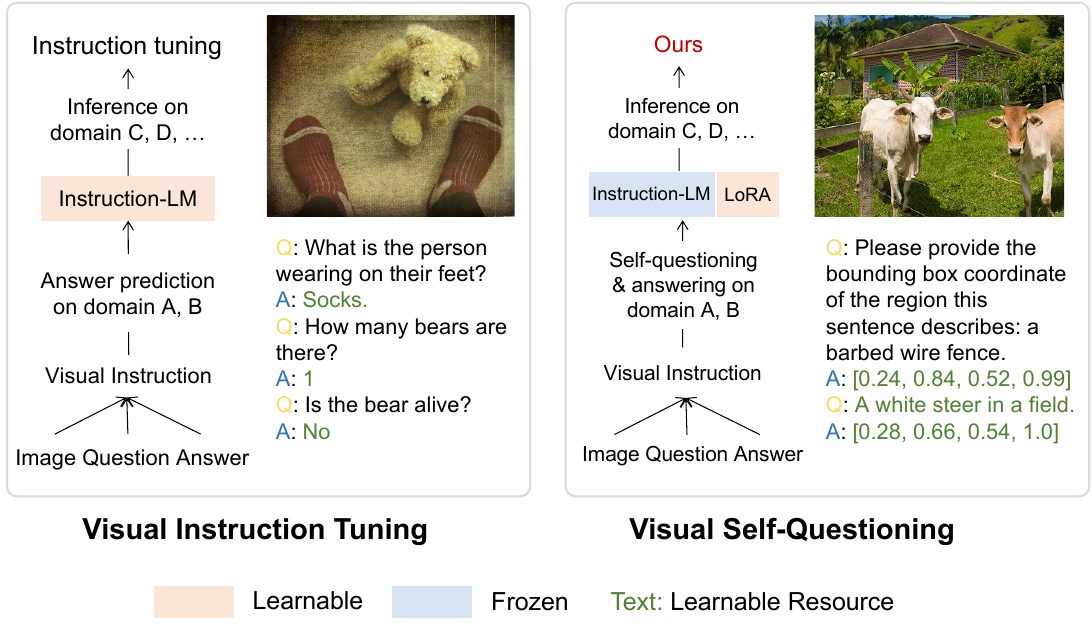}
  \caption{\label{fig:cover1}Training strategy comparison}
\end{subfigure}%
\begin{subfigure}{.33\textwidth}
  \centering
  \includegraphics[width=\linewidth]{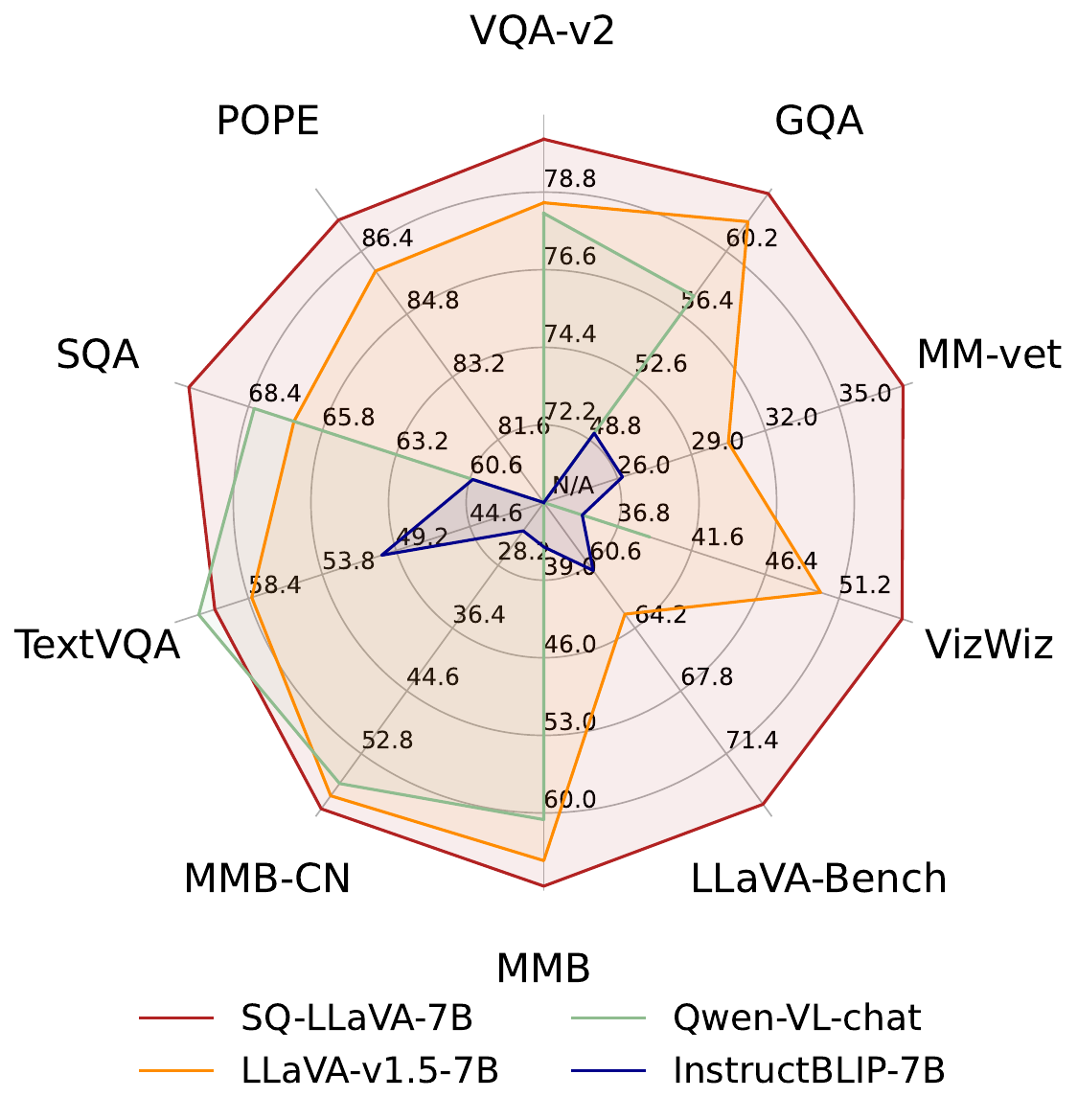}
  \caption{\label{fig:cover2}Performance comparison}
\end{subfigure}
\caption{ \label{fig:coverimage} (a) Comparision between visual instruction tuning and visual self-questioning (ours) for vision-language assistant. (b) The proposed SQ-LLaVA achieves state-of-the-art performance on 9 out of 10 tasks compared with other open-ended models.}
\end{figure}

LLaVA~\cite{llava1-5} model family usually consists of a pre-trained vision encoder (e.g., CLIP-ViT~\cite{clip}), a large generative language model (LLMs like Vicuna~\cite{vicuna}, LLaMA~\cite{zhang2023llama}, QWen~\cite{Bai2023QwenVLAV}, etc.), and a vision-to-language projector implemented by a few linear layers. 
However, the modality gap between the pre-trained vision encoder and the language model restricts both sides' generalization ability and feature representation. To overcome this challenge, various techniques have been proposed to align the vision and language domains, which could be roughly categorized into three groups: 1) build a more robust image feature extractor ~\cite{zhang2023llama,zhu2023minigpt,Dai2023InstructBLIPTG}, 2) collect more high-quality training data~\cite{llava1-5, Chen2023ShareGPT4VIL,zhu2023minigpt,Dai2023InstructBLIPTG}, and 3) fully fine-tune the vision and language models simultaneously during the pre-training stage~\cite{Chen2023ShareGPT4VIL}. While these methods have shown good progress in mitigating the domain gap, they inevitably bring higher computational costs and more expensive data collection and may also require sophisticated manual designs as well as heavy annotating efforts. Moreover, images generally encompass rich information, including color, context, and the relationships between objects, but most existing visual instruction datasets capture only a fraction. We posit that leveraging such under-explored knowledge could significantly aid in vision-language understanding. In this study, we propose a new \emph{visual self-questioning} approach (Fig.~\ref{fig:cover1}) by training the LLM to ask questions and discover vision clues without collecting extra data from other sources. 

\begin{figure}[t]
\centering
\begin{subfigure}{.5\textwidth}
  \centering
  \includegraphics[width=0.97\textwidth]{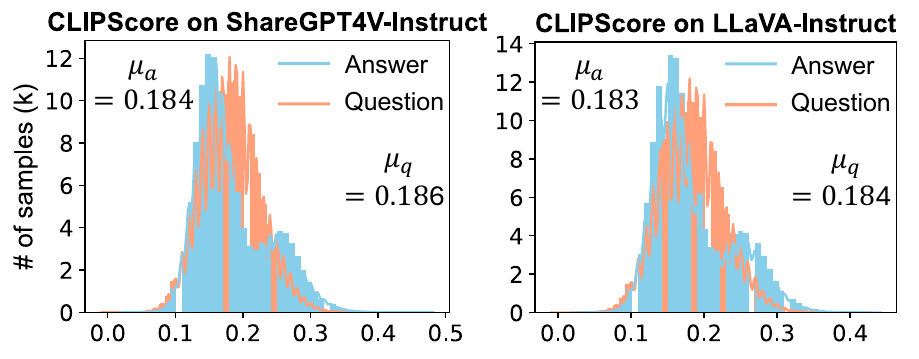}
\end{subfigure}%
\begin{subfigure}{.5\textwidth}
  \centering
  \includegraphics[width=0.97\textwidth]{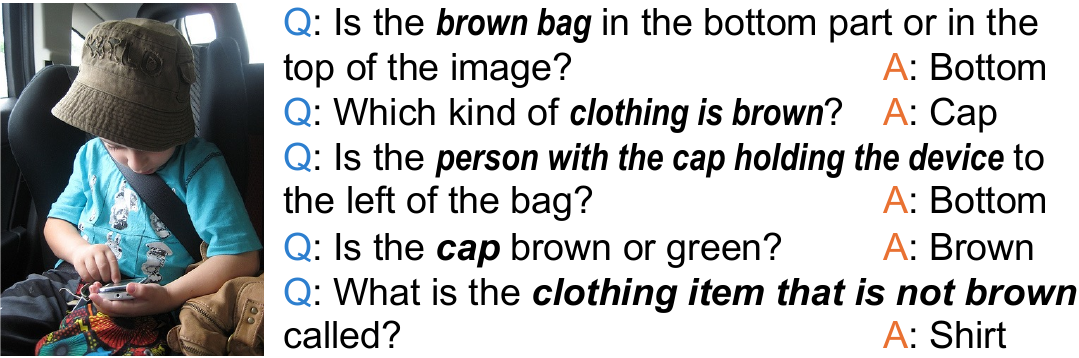}
  \vspace{2mm}
\end{subfigure}
\caption{\textbf{\emph{Left}}: Questions spread more samples around a higher mean CLIPScore than answers. \textbf{\emph{Right}}: Example of highly image-relevant questions within the visual instructional dataset for training a Vision-Language assistant.}\label{fig:motivation}

\end{figure}

Unlike existing visual instruction tuning methods that focus solely on answer prediction, the proposed visual self-questioning aims to extract relevant question context. As illustrated in Fig.~\ref{fig:motivation} \emph{right}, the questions could contain more image-related information than the answers, inspiring us that aligning images and related questions may further improve the model's vision-language understanding capacity. Quantitatively, we compute the CLIPScore~\cite{clip} (a higher value means better visual-text relevance) for all the image-question/-answer pairs on two visual instruction datasets (\texttt{LLaVA-instruct}~\cite{llava1-5} and \texttt{ShareGPT4V-instruct}~\cite{Chen2023ShareGPT4VIL}). By comparison, the questions' mean CLIPScore $\mu_q$=$0.184$ is larger than the answers' $\mu_a$=$0.183$ on \texttt{LLaVA-instruct}; $\mu_q$=$0.186$ is larger than the answers' $\mu_a$=$0.184$ on \texttt{ShareGPT4V-instruct}.
Fig.~\ref{fig:motivation} \emph{left} shows the CLIPScore distribution, indicating questions have a similar and even better visual relevance than the answers.  Therefore, questions in current visual instruction data can be used to fine-tune instructional LLMs owing to their diverse semantic information.

This work introduces a self-questioning LLaVA, namely SQ-LLaVA, to fully utilize questions within the instruction data as an additional learning resource for training instructional LLMs and empowering the model's curiosity (questioning ability). To efficiently align the vision and language domains, we apply LoRAs\cite{hu2021lora} to optimize both the vision encoder and the instructional LLM within the SQ-LLaVA. Plus, we develop a prototype extractor to enhance visual representation by leveraging learned clusters with meaningful semantic information to improve vision-language alignment further. Extensive experiments demonstrate that SQ-LLaVA surpasses existing visual instruction tuning methods in general vision understanding (see Fig.~\ref{fig:cover2}).
We summarize the contributions as follows.
\begin{itemize}
    \setlength\itemsep{0em}
    \item We propose a novel training technique, visual self-questioning for vision-language assistants (SQ-LLaVA), by leveraging highly relevant question contexts in instruction data. This SQ learning task promotes instructional LLMs to understand the relationship between images and questions, enhancing vision-language alignment without needing new data collection.
    
    \item We design and develop a lightweight tuning architecture for SQ-LLaVA, consisting of ViT-LoRA, LLM-LoRA, and a prototype extractor.
    The prototype extractor enhances vision embeddings, while ViT-LoRA and LLM-LoRA efficiently align vision and language domains during training.

    \item Extensive experimental results show that the proposed SQ-LLaVA leads to better performance in several tasks, including visual question-answering, visual instruction benchmarks, and zero-shot image captioning.  
\end{itemize}

\section{Related Work}
\subsection{Instruction Tuning}
Instruction tuning emerged as a pivotal methodology within the realm of natural language processing (NLP), facilitating Large Language Models (LLMs) such as GPT-3~\cite{gpt3}, PaLM~\cite{Chowdhery2022PaLMSL}, and LLaMA~\cite{Touvron2023LLaMAOA} to interpret and execute human language instructions across a spectrum of NLP tasks. This approach diverges from traditional fine-tuning mechanisms by incorporating a specialized data structure, termed instruction-following data~\cite{Wei2021FinetunedLM}, which is instrumental in the fine-tuning process of LLMs. Generally, there are two main categories regarding instruction tuning methods -- 1) closed-domain and 2) open-domain. 
The closed-domain instruction tuning~\cite{Wei2021FinetunedLM,Sanh2021MultitaskPT,Xu2023WizardLMEL} studies engaged LLMs with a comprehensive assortment of publicly accessible datasets and subsequently assessed their performance across diverse NLP tasks~\cite{Triantafillou2019MetaDatasetAD}. The empirical evidence from these inquiries consistently indicated that integrating varied NLP task instructions significantly augments the LLMs' efficacy in navigating novel tasks. Nonetheless, LLMs calibrated with such closed-form instructions exhibited limitations in real-world user scenarios, prompting the development of an alternative approach.
To address these constraints, the concept of open-domain instruction tuning~\cite{Ouyang2022TrainingLM,Wang2022SuperNaturalInstructionsGV} is conceived. OpenAI pioneered this approach by employing human annotators to compile a corpus of real-world question-answer datasets. These datasets form the foundation for training a reward model through reinforcement learning methodologies. The trained reward model then functions as a supervisory mechanism for further training instruction-oriented language models, such as InstructGPT~\cite{instructGPT} and Vicuna~\cite{vicuna}. This innovation marks a significant advancement in the field, aiming to bridge the gap between LLM performance and real-world applicability by leveraging instruction data derived from authentic user interactions.

\subsection{Large Vision Language Models}
As the field of LLMs and instruction tuning undergoes rapid advancements, the academic research community is progressively focusing on integrating visual information into LLMs to facilitate visual instruction tuning. This emerging area of research has witnessed the development of various methodologies~\cite{llava,llava1-5, Chen2023ShareGPT4VIL,Bai2023QwenVLAV,zhang2023llama} based on foundational vision-language models~\cite{clip,Wang2024TMASS,Sun2024ELIP,Li2022BLIPBL} and diverse LLM architectures~\cite{vicuna, Touvron2023Llama2O, Bai2023QwenVLAV}. In particular, LLaVA~\cite{llava} pioneers the integration of an LLM with a CLIP vision encoder to construct a vision language model, demonstrating remarkable capabilities in image-text dialogue tasks through pretraining alignment strategies and targeted instruction tuning. Subsequent investigations have sought to refine visual instruction tuning by enhancing the quality and variety of the datasets used during the pre-training and fine-tuning phases. Building upon these advancements, recent studies like LLaVA-v1.5~\cite{llava1-5} and ShareGPT4V~\cite{Chen2023ShareGPT4VIL} have achieved notable success in general vision-language comprehension, showcasing their ability to undertake complex question-answering tasks. This progression underscores the importance of sophisticated data handling and model tuning strategies in developing effective vision-language models.


\begin{figure*}[t]
\centering
\includegraphics[width=\textwidth]{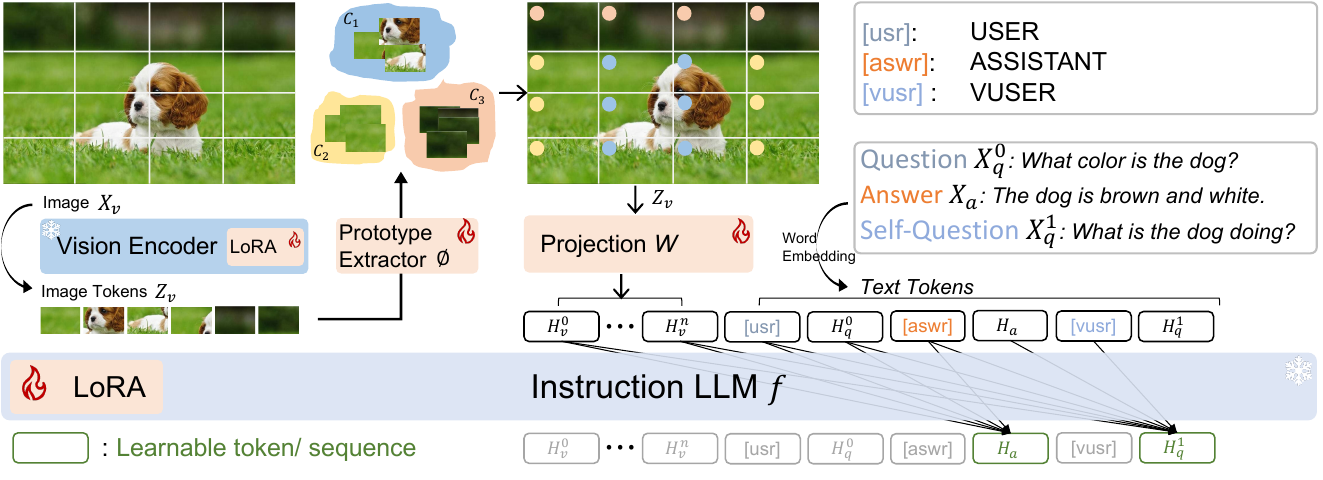}
  \caption{ \label{fig:framework}Model architecture of SQ-LLaVA. We propose prototype extractor to extract visual clustering information to enhance the visual embedding encoded by the visual encoder. SQ-LLaVA defines a new token \texttt{[\textcolor{Mycolor1}{vusr}]} as specific instruction for LLM to perform visual self-questioning. Besides question answering, SQ-LLaVA treats questioning as another training objective.   }
\end{figure*}

\section{Method }
\label{sec:methods}
\subsection{Architecture Overview}\label{architecture}
The proposed SQ-LLaVA model (see Fig.~\ref{fig:framework}) consists of four main components: 1) A pre-trained vision encoder CLIP-ViT~\cite{clip} that extracts a sequence embedding of image tokens $Z_v$ for an input image $X_v$; 2) A prototype extractor $\phi(\cdot)$ learning visual clusters to enhance the original image tokens; 3) A trainable projection block $W(\cdot)$ with two linear layers to map the enhanced image tokens to the language domain tokens $H_v$, handling the dimension misalignment between the vision and language domain; and 4) Our LLM backbone $f(\cdot)$ implemented by the pre-trained Vicuna~\cite{vicuna} to predict the next token upon the previous embedding sequence. 
Given the input question $X_q$ and answer $X_a$, a word embedding matrix is used to map them to contextual embeddings $H_q$ and $H_a$, and the distribution over $H_a^{(i+1)}$ can be obtained following the auto-regressive model as:
\begin{equation}\label{eq:1}
    p_\theta(H_a^{(i+1)} \mid H_v, H_q,H_a^{(1:i)})=\sigma(f(H_v,H_q,H_a^{(1:i)})),
\end{equation}
where $\theta$ represents all the trainable parameters in our model, $\sigma(\cdot)$ is a softmax function, and $f(\cdot)$ outputs the last token embedding of the whole sequence. We denote $p_\theta$ as the prediction probability for the anticipated answer token $H_a^{(i+1)}$ at the position $i+1$, conditioning on the input image token embedding $H_v$, the question token embedding $H_q$, and the previous answer token embeddings $H_a^{(1:i)}$. As shown in Eq.~\eqref{eq:1}, the proposed SQ-LLaVA applies the language model $f(\cdot)$ to model $p_\theta$ given by $H_v$, $H_q$, and $H_a^{(1:i)}$.
Existing visual instruction tuning methods~\cite{llava1-5,Chen2023ShareGPT4VIL} are only able to predict $H_a$, yet cannot fully exploit the rich semantic clues within $H_q$. 
In this study, we propose visual self-questioning instructions to guide LLMs in capturing the visual content within questions.

\subsection{Visual Self-questioning Instruction} 
In broad real-world scenarios, proactively asking a question requires more understanding and background knowledge than answering~\cite{science1}. Similarly, this work proposes visual self-questioning to encourage the LLM to discover deeper vision-language alignment and improve the overall instruction-following performance.
Particularly, SQ-LLaVA treats questioning as a new learning objective beyond answering, which, to the best of our knowledge, is the first practice in the field of visual instruction tuning. While the current vision-language model can ask questions~\cite{2023GPT4VisionSC}, such skill is still learned from question-answering through instruction tuning. However, our proposed method shows that the decoder-based LLM has the potential to learn more skills, such as how to ask questions spontaneously when a unique instruction token is given (e.g., we define [\textcolor{Mycolor1}{vusr}] in our work). Furthermore, visual self-questioning can potentially improve general vision-language understanding. To be specific, as shown in Fig.~\ref{fig:coverimage}, there are a certain amount of questions containing more meaningful image-related information than answers in the existing visual instruction data~\cite{llava1-5,Chen2023ShareGPT4VIL}. Thus, we hypothesize that the vision-language understanding can be improved once the LLM learns how to predict relevant questions about a given image. 

\begin{figure}[t]
\centering
\includegraphics[width=0.9\textwidth]{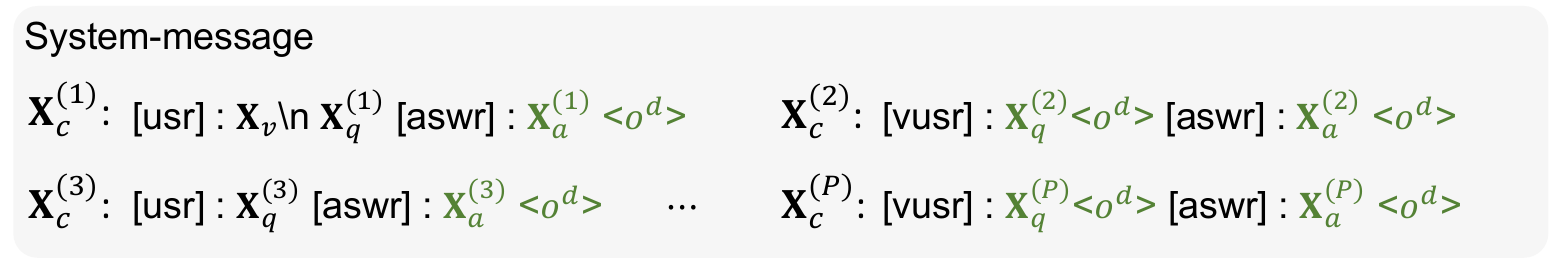} 
  \caption{ \label{fig:inputseq}The input sequence used to train SQ-LLaVA. Our model is trained to predict \emph{question}, \emph{answer}, and \emph{where to stop}. We use {\color{cGreen}tokens} to represent learnable tokens, where {\color{cGreen}$X_q$} is the question, {\color{cGreen}$X_a$} is the answer, and {\color{cGreen}$<o^{d}>$} is the delimiter token. In SQ-LLaVA, the \emph{System-message} = \emph{"The assistant gives helpful, detailed, and polite answers to the user's questions. Also, the assistant is a curious virtual user can ask complex questions that are relevant to the content in the image."}}
\end{figure}

\textbf{Self-Questioning Prompt Design}.
We provide ground-truth content for the visual self-questioning, restricting SQ-LLaVA from asking image-related questions. To this end, we leverage questions as another learnable resource and follow the regular auto-regressive training objective. 
As shown in Fig.~\ref{fig:inputseq}, the training data for SQ-LLaVA is designed in a format with a pre-defined template. To be specific, the \emph{system message} as a fixed prompt is added at the beginning of each instruction data, indicating a general job description (e.g., gives helpful answers and asks complex questions) for the LLM. Existing visual instruction tuning methods utilize the unique tokens (\texttt{[usr]}, \texttt{[aswr]}) to give the LLM a particular instruction (e.g., questioning understanding, answer prediction, etc.) and apply the delimiter token $<o^{d}>$ to mark the ending position. In this work, we propose a new special token \texttt{[\textcolor{Mycolor1}{vusr}]}, indicating a specific instruction -- asking questions. Combining with the delimiter, we can construct instructions for self-questioning as a new training task. 

Each sample of current visual instruction data consists of one image $X_v$ and $P$ question-answer pairs $(X_{q}^{(1)},X_{a}^{(1)}, \cdots ,X_{q}^{(P)},X_{a}^{(P)})$. We collect one question $X_q^{(j)}$ and its answer $X_a^{(j)}$ with special tokens to construct the $j^{th}$ turn conversation as
\begin{equation}\label{instructiondata}
    X_{c}^j=
    \left\{\begin{matrix}
     (\texttt{[usr]},X_{q}^{(j)}, \texttt{[aswr]},X_a^{(j)}) & ~~j=1 \mathrm{~or~} j>1,R< \delta\\ 
     ~(\texttt{[\textcolor{Mycolor1}{vusr}]},X_{q}^{(j)},\texttt{[aswr]},X_a^{(j)}) & j>1,R> \delta 
    \end{matrix}\right. ,
\end{equation}
where $R\in[0,1]$ is a random number and $\delta = 0.5$ is a threshold that sets the proportion of self-questioning pairs in the conversations. Finally, the full-text input $X_{c}$ will be mapped to textual embedding $H_{c}$ through word embedding.

SQ-LLaVA performs zero-shot questioning without any in-context knowledge or human language instruction since the only instructional prompt is a unique token \texttt{[\textcolor{Mycolor1}{vusr}]}. After visual self-questioning on instruction data with various question formats $X_q$, the questions sampled by SQ-LLaVA are more diversified than GPT4-V (as shown in Fig.~\ref{fig:self-q_exs}), since the LLM has learned alignment between image and questions, which is different from previous works~\cite{2023GPT4VisionSC,Wang2022SelfInstructAL}. Specifically, previous general-purpose vision language models such as GPT4-V~\cite{2023GPT4VisionSC} can generate questions based on a given image, but it requires explicit language instruction such as ``Ask complex questions that are relevant to the content in the image''. Accordingly, the quality of generated questions highly relies on prompt engineering. Also, self-instruct~\cite{Wang2022SelfInstructAL} utilizes in-context learning to prompt the LLM to ask specific questions.

\subsection{Enhanced Visual Representation} \label{proto}
Unlike previous visual instruction tuning methods, SQ-LLaVA jointly benefits from visual self-questioning and question-answering. For better visual self-questioning, we develop a prototype extractor that recognizes and groups similar patterns of visual information from the latent space. Our primary goal is to enhance visual representation through prototype learning. 

Specifically, we utilize clustering to extract centroid representations of image tokens $Z_v$, where each cluster center is treated as a prototype, which, in return, will be distributed to each of the original image token embeddings. Our proposed prototype extractor $\phi(\cdot)$ is a lightweight design involving two parts: 1) cluster center optimization and 2) prototype information distribution. Following~\cite{cluster1,cluster2}, we randomly initialize $K=256$ cluster centers $C$ and deploy the iterative Expectation-Maximization (EM) clustering process to capture representative semantics in the latent space by 
\begin{equation}
    \begin{aligned}
        \text{E-step}:~~&  \mathcal{M}^{(t)}=\sigma(q(C^{(t)})*k(Z_v)^\top), \\ \text{M-step}:~~& C^{(t+1)}=\mathcal{M}^{(t)}*v(Z_v),
    \end{aligned}
\end{equation}\label{eq:EM}where $\mathcal{M}^{(t)} \in \left[ 0,1 \right]$ denotes a soft cluster assignment map at the $t^{th}$ iteration, $\sigma (\cdot)$ is a softmax function, and $t\in\left\{ 1,\cdots ,T \right\}$ indexes the iteration of EM step with $T=2$ in this work. Three trainable linear layers $q$, $k$, and $v$ are used in~\eqref{eq:EM}, where $q(\cdot)$ projects the prototype $C$ to a query vector, and $k(\cdot)$ and $v(\cdot)$ project $H_v$ into key and value vectors, followed by a normalization layer, respectively. The prototype extractor iteratively updates cluster map $\mathcal{M}$ and centers $C$. 

After the prototype extraction, we train a linear layer $z(\cdot)$ to adaptively map the visual cluster information to the raw image embedding $Z_v$. For the $i^{th}$ token embedding $Z_v^{(i)}$, we update it as
\begin{equation}\label{eq:hv_proto}
    Z_v^{(i)}=Z_v^{(i)}+z(\frac{1}{K}\sum\nolimits_{j=1}^{K}S_{c}(C_j,Z_v^{(i)}) \times C_j),
\end{equation}
where $S_c (\cdot,\cdot)$ is a normalized cosine similarity function. The weighted sum over prototypes in Eq.~\eqref{eq:hv_proto} emerges as an indispensable step for contextual understanding from image tokens, recognizing and grouping similar patterns and semantics. It clusters image tokens as prototypes that display homogeneity in semantics, such as ``grass'' and ``dog''. The prototypes can describe the intrinsic semantic meanings by aggregating entities that exhibit shared attributes. Finally, we map the image sequence embedding $Z_v$ to language domain $H_v$ with a two-layer linear projector $W(\cdot)$.

\subsection{Model Training}
\subsubsection{Stage1: Pre-training for Vision-Language Alignment.}
Unlike text-only LLMs, the vision-language model also fine-tunes LLM using image tokens as input (see Fig.~\ref{fig:framework}). Therefore, the pre-training stage aims to optimize the LLM by explicitly executing the visual instruction. The proposed SQ-LLaVA adopts Vicuna~\cite{vicuna} as its instruction LLM, pre-training on massive text corpora to predict the next text token given the previous context, not only containing text but also visual instructions. To achieve this, we organize the pre-training data as $D_{PT}=\{ [X^{(1)}_v,X^{(1)}_a] ,\cdots, [X^{(N)}_v,X^{(N)}_a] \}$, where $N$ is the total number of training samples, and each sample has an image and its related descriptions. Each image and text input pair will be mapped to sequence embeddings ($H_v$ and $H_a$) as elaborated in Section~\ref{architecture}. 
During pre-training, we freeze the vision encoder and LLM and mainly train the prototype extractor $\phi$ and the vision-to-language projector $W$. The pre-training goal is to maximize the probability of the predicted image description $H_a$ given an image $H_v$. When training a visual instructional LLM, we follow the negative log-likelihood objective function as 
\begin{equation}\label{eq:objective}
    \sum_{v,a\in D_{PT}}^{}-\mathrm{log}p_\theta (H_a\mid H_v)=\sum_{v,a\in D_{PT}}^{}\sum_{i=1}^{L}-\mathrm{log}p_\theta(H_a^{(i+1)}\mid H_v,H_a^{(1:i)}),
\end{equation}
where $L$ denotes the sequence length of answer tokens in $H_a$, $\theta$ is the total trainable parameter of $\phi$ and $W$, $p(H_a\mid H_v)$ can be computed by Eq.~\eqref{eq:1}, and $H_a^{(1:i)}$ represents all the answer tokens before the current prediction $H_a^{(i+1)}$.

\subsubsection{Stage2: Fine-tuning.}
Existing methods, such as LLaVA~\cite{llava,Chen2023ShareGPT4VIL}, mainly update the vision-to-language projector (usually a couple of linear layers) and the language model during fine-tuning. Nevertheless, the projector might be too weak to capture the relationship between the image and the questions. 
Following the previous multi-modal understanding method~\cite{clip}, we unfreeze the vision encoder and LLM during fine-tuning for a joint optimization further to eliminate the gap between the vision and language domain. 

To mitigate the heavy computational overhead, we take advantage of LoRA~\cite{hu2021lora} as a lightweight tuning option that can achieve similar (even better) performance to fully fine-tuning when training large models on a relatively small dataset. We add LoRA in both the vision encoder and LLM. Thus, the learnable parameters $\theta$ of the proposed SQ-LLaVA during fine-tuning represent a combination of all the parameters of LLM-LoRA, ViT-LoRA, prototype extractor $\phi$, and the vision-to-language projector $W$. 
Given the instruction tuning data $D_{IT}=\{ [X^{(1)}_v,X^{(1)}_{c}] ,\cdots, [X^{(N)}_v,X^{(N)}_{c}] \}$, we take the conversational data $X_c$ and the image $X_v$ as input, mapping them to sequence embedding ($H_c$ and $H_v$) as elaborated in Section~\ref{architecture}, and minimize the negative log-likelihood loss for the \emph{self-questioning} and \emph{answering} tasks as follows
\begin{align}
    \text{Self-questioning}:&~~\sum_{v,c\in D_{IT}}^{}-\mathrm{log} p_\theta(H_q^{(j+1)} \mid H_v,H_{c}^{(1:j)})\label{eq:selfq-obj}, \\
    \text{Answering}:&~~\sum_{v,c\in D_{IT}}^{}-\mathrm{log}p_\theta(H_a^{(j+1)} \mid H_v,H_{c}^{(1:j)},H_q^{(j+1)})\label{eq:qa-obj},
\end{align}
where $j\in\left\{ 1,\cdots ,P \right\}$, indicating the index of question or answer within the conversational data $X_c^{(*)}$.
Notably, previous works~\cite{llava1-5,Chen2023ShareGPT4VIL,zhu2023minigpt} only involve answering tasks, but we introduce visual self-questioning as an additional training task for visual instruction tuning. 
Eventually, SQ-LLaVA, as a vision-language assistant, not only executes human instructions by optimizing the objective function in Eq.~\eqref{eq:qa-obj} but can raise questions out of the given image after optimizing the Eq.~\eqref{eq:selfq-obj}. This capability potentially yields more diverse question-answer guidance and enhances multi-modal understanding.

\section{Experiments}
\label{sec:experiments}

\subsection{Experimental Setting}
\subsubsection{Dataset.}
Our work uses the open-source visual instruction dataset provided by LLaVA~\cite{llava1-5} and ShareGPT4V~\cite{Chen2023ShareGPT4VIL} for training. Each dataset has large-scale image-text paired data for pre-training and instruction-following data for fine-tuning. Specifically, the instruction data proposed by LLaVA is a comprehensive mixture of COCO, GQA~\cite{gqa}, OCR-VQA~\cite{ocrvqa}, TextVQA~\cite{textvqa}, VisualGenome~\cite{Krishna2016VisualGC}, RefCOCO~\cite{refcoco}, and image from InstructBLIP~\cite{Dai2023InstructBLIPTG} involves multiple reasoning, spatial understanding, multi-step inference, optical character recognition, and grounding of visual concepts to language. The ShareGPT4V dataset consists of the same mixture image as LLaVA but enrolls more images from other datasets such as SAM~\cite{Kirillov2023SegmentA} and  WebData~\cite{webdata1,laion}. 

We evaluate our model on general vision-language understanding tasks. Specifically, we use ten visual oriented question answering benchmarks: VQA$^{v2}$~\cite{vqav2}; GQA~\cite{gqa}; VizWiz~\cite{Gurari2018VizWizGC}; SQA$^I$~\cite{learnToexplain}: ScienceQA-IMG; VQA$^T$: TextVQA~\cite{textvqa}; POPE~\cite{pope}; MM-Vet~\cite{Yu2023MMVetEL}; LLaVA$^W$~\cite{llava}: LLaVA (in the wild); MMB: MMBenchmark~\cite{Liu2023MMBenchIY}; MMB$^{CN}$: MMBench-Chinese~\cite{Liu2023MMBenchIY}. We report the prediction accuracy on all the benchmarks.
We also evaluate our model for visual information discovery through captioning. To be specific, we employ testing images from four datasets, \emph{i.e.}, COCO~\cite{coco}, Flickr30K~\cite{flickr}, Conceptual~\cite{sharma-etal-2018-conceptual}, and Nocaps~\cite{agrawal2019nocaps}. 
Following~\cite{Dess2023CrossDomainIC}, we evaluate the proposed method with regular image captioning metrics, \emph{e.g.}, BLEU~\cite{bleu} and CIDEr~\cite{Vedantam2014CIDErCI}. For the open-world methods LLaVA-v1.5~\cite{llava1-5}, ShareGPT4V~\cite{Chen2023ShareGPT4VIL} and the proposed SQ-LLaVA, we utilize greedy search for caption generation with a prompt of ``Provide a brief description of the given image, your answer should be in one sentence.'' The generated caption is used to evaluate the performance of image captioning.

\subsection{Implementation}
We adopt Vicuna~\cite{vicuna} as the pre-trained language generative model and CLIP-ViT~\cite{clip} as the vision encoder. We pre-train the prototype extractor and the vision-to-language projector using AdamW~\cite{adamw} optimizer with a learning rate of $2\times10^{-3}$ and a constant scheduler for one epoch. Following previous work~\cite{llava1-5}, we keep the global batch size as $256$ for pre-training and $128$ for fine-tuning. During fine-tuning, we insert LoRA\cite{hu2021lora} with $rank = 128$ and $\alpha = 256$ into the language model (LLM-LoRA) and LoRA with $rank = 32$ and $\alpha = 64$ into the vision encoder (ViT-LoRA). We optimize LoRA modules, the prototype extractor, and vision-to-language projector for one epoch. We set the learning rate to $2\times10^{-4}$ for LoRA, and $2\times10^{-5}$ for the other layers. 
All the weights of the pre-trained language model and vision encoder remain fixed during fine-tuning.

\begin{table*}[t]
\centering
\tabcolsep=0.22em
\caption{\label{tab:vqa} Comparison with state-of-the-art methods on ten benchmarks. After training on the same instruction data, SQ-LLaVA traind on~\cite{llava1-5} and SQ-LLaVA$^*$ trained on~\cite{Chen2023ShareGPT4VIL} surpass their baseline model LLaVA-v1.5 and ShareGPT4V on 9 out of 10 and 6 out of 10 benchmarks in the 7B scale, and 8 out of 10 and 6 out of 10 in the 13B scale. The best results are \textbf{bold} and the second-best results are \underline{underlined}. }
\scalebox{0.8}{
\begin{tabular}{ll|cccccccccc}
\toprule
LLM& Model & VQA$^{v2}$ & GQA & VizWiz & SQA$^I$ & VQA$^T$ & POPE&MM-Vet&LLaVA$^W$&MMB&MMB$^{CN}$ \\ \midrule
\multirow{7}{*}{7B}&InstructBLIP~\cite{Dai2023InstructBLIPTG}&-&49.2&34.5&60.5&50.1&-&26.2&60.9&36.0&23.7\\
&Qwen-VL~\cite{Bai2023QwenVLAV}&78.8&59.3&35.2&67.1&\textbf{63.8}&-&-&-&38.2&7.4\\
&Qwen-VL-chat~\cite{Bai2023QwenVLAV}&78.2&57.5&38.9&68.2&\underline{61.5}&-&-&-&60.6&56.7\\
&LLaVA-v1.5~\cite{llava1-5} &78.5 &62.0&50.0& 66.8&58.2&85.9&30.5&63.4&64.3&58.3   \\
&ShareGPT4V~\cite{Chen2023ShareGPT4VIL} &\textbf{80.6} &\underline{63.3}&\textbf{57.2}& 68.4&60.4&86.8&\textbf{37.6}&\underline{72.6}&\textbf{68.8}&\textbf{62.2}   \\ 
 &\cellcolor{cGrey}SQ-LLaVA &\cellcolor{cGrey}79.2&\cellcolor{cGrey}62.8&\cellcolor{cGrey}54.0&\cellcolor{cGrey}\underline{68.9}&\cellcolor{cGrey}58.6&\cellcolor{cGrey}\textbf{87.7}&\cellcolor{cGrey}32.5&\cellcolor{cGrey}66.3&\cellcolor{cGrey}66.2&\cellcolor{cGrey}58.1\\
 &\cellcolor{cGrey}SQ-LLaVA$^*$ &\cellcolor{cGrey}\underline{80.3}&\cellcolor{cGrey}\textbf{63.7}&\cellcolor{cGrey}\underline{55.3}&\cellcolor{cGrey}\textbf{70.5}&\cellcolor{cGrey}60.5&\cellcolor{cGrey}\underline{87.2}&\cellcolor{cGrey}\textbf{37.6}&\cellcolor{cGrey}\textbf{74.3}&\cellcolor{cGrey}\underline{66.6}&\cellcolor{cGrey}\underline{60.0}\\ \midrule
 \multirow{5}{*}{13B}&InstructBLIP~\cite{Dai2023InstructBLIPTG}&-&49.5&33.4&63.1&50.7&78.9&25.6&58.2&-&-\\
& LLaVA-v1.5~\cite{llava1-5} &80.0 &63.3&53.6& \textbf{71.6}&58.2&85.9&35.4&70.7&67.7&\underline{63.6}   \\
 &ShareGPT4V~\cite{Chen2023ShareGPT4VIL} &\underline{81.0}&\underline{64.8}&\underline{55.6}&71.2&\textbf{62.2}&-&\textbf{43.1}&\underline{79.9}&\underline{68.5}&\textbf{63.7} \\
  &\cellcolor{cGrey}SQ-LLaVA &\cellcolor{cGrey}80.1&\cellcolor{cGrey}63.6&\cellcolor{cGrey}54.6&\cellcolor{cGrey}69.8&\cellcolor{cGrey}60.2&\cellcolor{cGrey}\textbf{87.7}&\cellcolor{cGrey}35.5&\cellcolor{cGrey}74.6&\cellcolor{cGrey}\textbf{68.7}&\cellcolor{cGrey}62.0\\
& \cellcolor{cGrey}SQ-LLaVA$^*$ &\cellcolor{cGrey}\textbf{81.3}&\cellcolor{cGrey}\textbf{65.0}&\cellcolor{cGrey}\textbf{58.2}&\cellcolor{cGrey}\underline{71.5}&\cellcolor{cGrey}\underline{61.9}&\cellcolor{cGrey}\underline{87.4}&\cellcolor{cGrey}\underline{39.7}&\cellcolor{cGrey}\textbf{80.7}&\cellcolor{cGrey}\underline{68.5}&\cellcolor{cGrey}62.5\\

 \bottomrule
\end{tabular}}
\end{table*}

\begin{figure*}[t]
\centering
\includegraphics[width=0.98\textwidth]{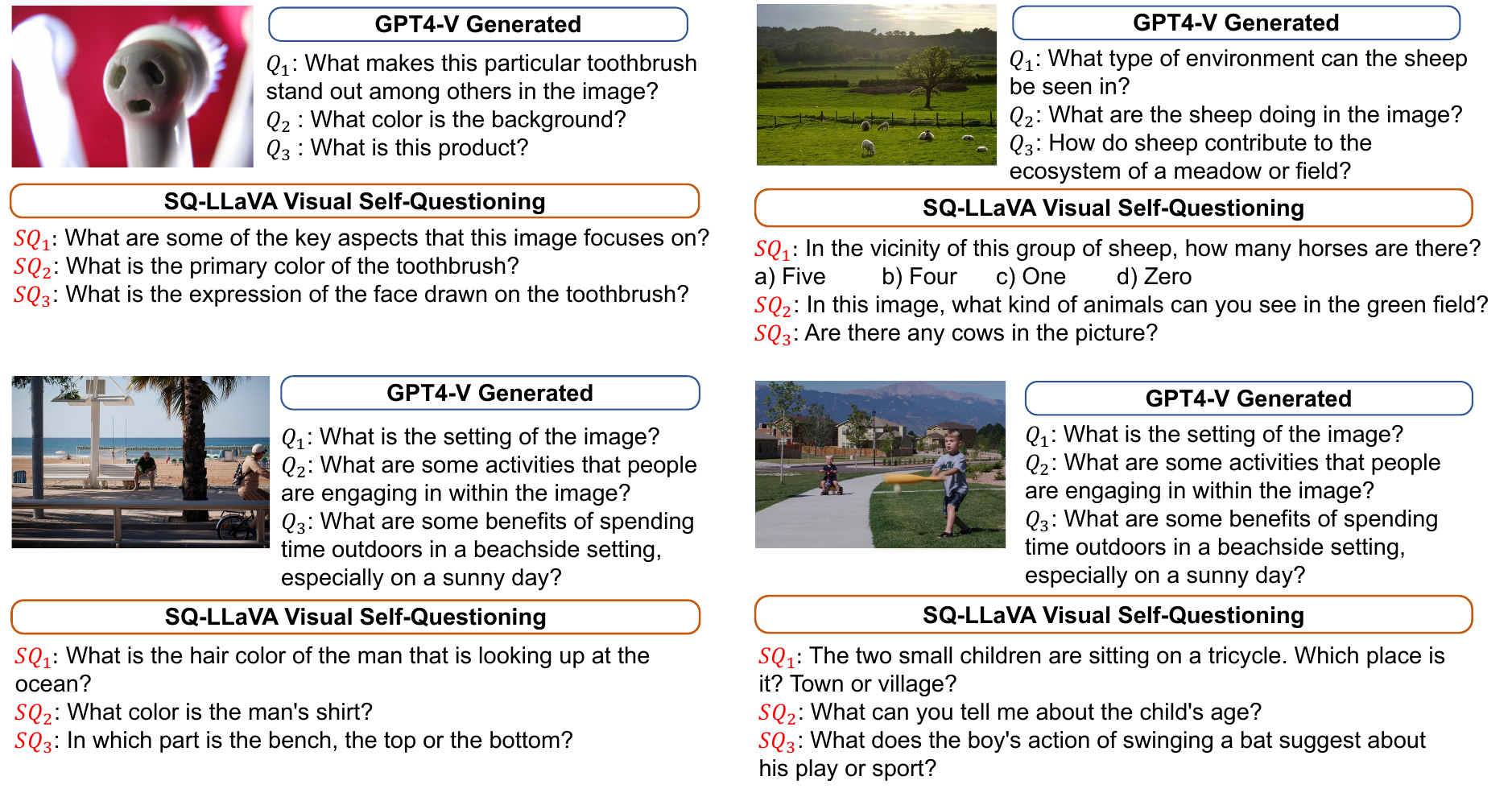}
  \caption{ \label{fig:self-q_exs} Visual self-questioning of SQ-LLaVA-7B. Comparing to the question data provided by GPT4-V (data collected by LLaVA-v1.5~\cite{llava1-5}), SQ-LLaVA can generate questions with higher diversity \emph{i.e.} multiple choice and tricky questions. }
\end{figure*}

\subsection{Zero-shot Multilingual Capability}

We evaluate SQ-LLaVA on ten benchmarks, covering a range of academic Visual Question Answering (VQA) tasks and recent instruction tuning tasks designed for large vision language models. The academic VQA includes VQA-v2~\cite{vqav2} and VizWiz~\cite{Gurari2018VizWizGC}. GQA~\cite{gqa} is a fine-grained real-world visual reasoning and question-answering benchmark. ScienceQA~\cite{learnToexplain} is a benchmark with rich subjects (natural science, language science, and social science). TextVQA~\cite{textvqa} requires the model to recognize the texts in the image. LLaVA (in the wild) and MM-Vet~\cite{Yu2023MMVetEL} use GPT4 to assess the capability of the models for testing. With manually designed questions, MM-Bench and MMBench-CN~\cite{Liu2023MMBenchIY} evaluate the model's vision-related reasoning and perception for English and Chinese, respectively. POPE~\cite{pope} is a benchmark for evaluating object hallucination~\cite{rohrbach-etal-2018-object}.

In Table~\ref{tab:vqa}, we quantitatively compare between SQ-LLaVA and existing models. SQ-LLaVA-7B and SQ-LLaVA-13B trained on two instruction datasets~\cite{llava1-5, Chen2023ShareGPT4VIL} outperform previous methods in six out of ten visual instruction tuning tasks. 
To be specific, SQ-LLaVA-7B achieves $17.2 \%$ improvement over LLaVA-v1.5-7B on the LLaVA (in the wild) benchmark, indicating the superior capabilities of our model in tasks such as detailed description and complex reasoning. Also, SQ-LLavA-7B improves over previous methods on ScienceQA, indicating that our model excels in understanding and reasoning over scientific content and can effectively handle multi-modal information. The improvement in ScienceQA suggests strong capabilities in multi-hop reasoning, comprehension of complex scientific concepts, and the ability to utilize context and explanations to derive correct answers. SQ-LLaVA-7B has a steady improvement over LLaVA-v1.5-7B and ShareGPT4V-7B on the POPE benchmark, and the $2\%$ and $1\%$ improvement indicates that our proposed method has better reliability and trustworthiness since POPE is a task designed to evaluate the phenomenon of object hallucination~\cite{pope,rohrbach-etal-2018-object}.
In the bottom section of Table~\ref{tab:vqa}, the proposed SQ-LLaVA-13B surpasses previous works in six out of ten benchmarks, indicating the scalability of our method on larger LLM.
Notably, the performance inconsistency on some datasets might be due to the unsupervised prototype extractor (lacking pixel-wise guidance) in our model. To mitigate this issue, we could leverage pseudo object masks (e.g., given by the pre-trained segment anything) in learning prototypes. Despite this limitation, all the improvements are achieved with significantly fewer trainable parameters compared to other methods~\cite{Chen2023ShareGPT4VIL,llava1-5,Bai2023QwenVLAV}. 

\begin{table}[t]
\centering
\caption{ Comparison of image captioning on four datasets. SQ-LLaVA is trained on data collected by~\cite{llava1-5}. SQ-LLaVA$^*$ is trained on data collected by~\cite{Chen2023ShareGPT4VIL}.} 
  \begin{subtable}{.53\linewidth}
    \centering
    \scalebox{0.8}{
    \begin{tabular}{l|cccccc}
    \toprule
    \multirow{2}{*}{Model} & \multicolumn{2}{c}{Flickr30k}& \multicolumn{2}{c}{Nocaps$^{out}$} & \multicolumn{2}{c}{Conceptual} \\
    \cmidrule(lr){2-3}
    \cmidrule(lr){4-5}
    \cmidrule(lr){6-7}
     &  B@4 & CIDEr &B@4 & CIDEr & B@4 & CIDEr   \\ 
    \midrule
    ClipCap~\cite{Mokady2021ClipCapCP} & 17.21 & 41.65 & 20.32 & 51.74 & 1.47 & 23.74 \\
    DiscriTune~\cite{Dess2023CrossDomainIC} & 18.48&44.78&24.10 & 57.06 & 1.71 & 28.01 \\
    LLaVA-v1.5\cite{llava1-5} & 28.67 & 81.27 & 35.78 & 103.56 & 2.79 & 39.20 \\
    ShareGPT4V~\cite{Chen2023ShareGPT4VIL} &31.00&\textbf{86.17}& 37.19&\textbf{107.45}&2.78&37.86\\
    \rowcolor{cGrey}SQ-LLaVA  &29.88&83.51&36.21&105.42&2.90&39.49 \\
    \rowcolor{cGrey}SQ-LLaVA$^*$  &\textbf{31.49}&83.14&\textbf{37.20}&107.42&\textbf{2.91}&\textbf{41.24}\\
    \bottomrule
    \end{tabular}}
    \caption{\label{tab:cap1} Zero-shot image captioning}
  \end{subtable}%
  \begin{subtable}{.47\linewidth}
    \centering

    \scalebox{0.8}{
    \begin{tabular}{l|cc|cc}
    \toprule
    \multirow{3}{*}{Model}& \multicolumn{2}{c}{Zero-shot} & \multicolumn{2}{c}{Fine-tune} \\
    \cmidrule(lr){2-3}
    \cmidrule(lr){4-5}
     & B@4 & CIDEr & B@4 & CIDEr \\ \midrule
    ClipCap & 8.50 & 37.03 & 32.60 & 108.55 \\
    DiscriTune   & 13.99 & 53.20 & 32.31 & 105.40 \\
    eP-ALM~\cite{Shukor2023ePALMEP}  & 29.47 & 97.22 & 33.35 & 111.63 \\
    MAPL~\cite{Maas2022MAPLPA} & 12.30 & 54.30 & 36.45 & 125.20 \\
    LLaVA-v1.5 & \textbf{29.96}&\textbf{111.46}&-&-\\
    \rowcolor{cGrey}SQ-LLaVA & 29.85 & 110.77 & \textbf{40.76} & \textbf{136.78} \\
    \bottomrule
    \end{tabular}}
     \caption{\label{tab:cap2} Image captioning on COCO}
  \end{subtable}
\end{table}

\subsection{Visual Information Discovery} 
In this experiment, we showcase the diversity and reliability of the proposed SQ-LLaVA through various qualitative applications, including detailed image description, visual information summary, and visual self-questioning. We also present quantitative results on the task of image captioning.

\noindent\textbf{Abilities of SQ-LLaVA Through Qualitative Samples.} SQ-LLaVA exhibits numerous advanced capabilities compared to traditional vision-language models. Notably, SQ-LLaVA effectively mitigates object hallucination~\cite{pope,rohrbach-etal-2018-object}, resulting in predictions that are more trustworthy. Additionally, SQ-LLaVA is capable of generating diverse and meaningful questions about a given image without requiring human textual instructions. As illustrated in Fig.~\ref{fig:self-q_exs}, the generated questions include multiple-choice, reasoning, binary choice questions, and the content within each question, all of which are either highly correlated or tricky to the given image, such as ``What is the hair color of the man that is looking up at the ocean?'' from the third image and ``Are there cows in the picture?'' from the second image. These examples highlight SQ-LLaVA's ability to produce questions with high diversity and quality.

\begin{figure}[t]
\centering
\includegraphics[width=0.98\textwidth]{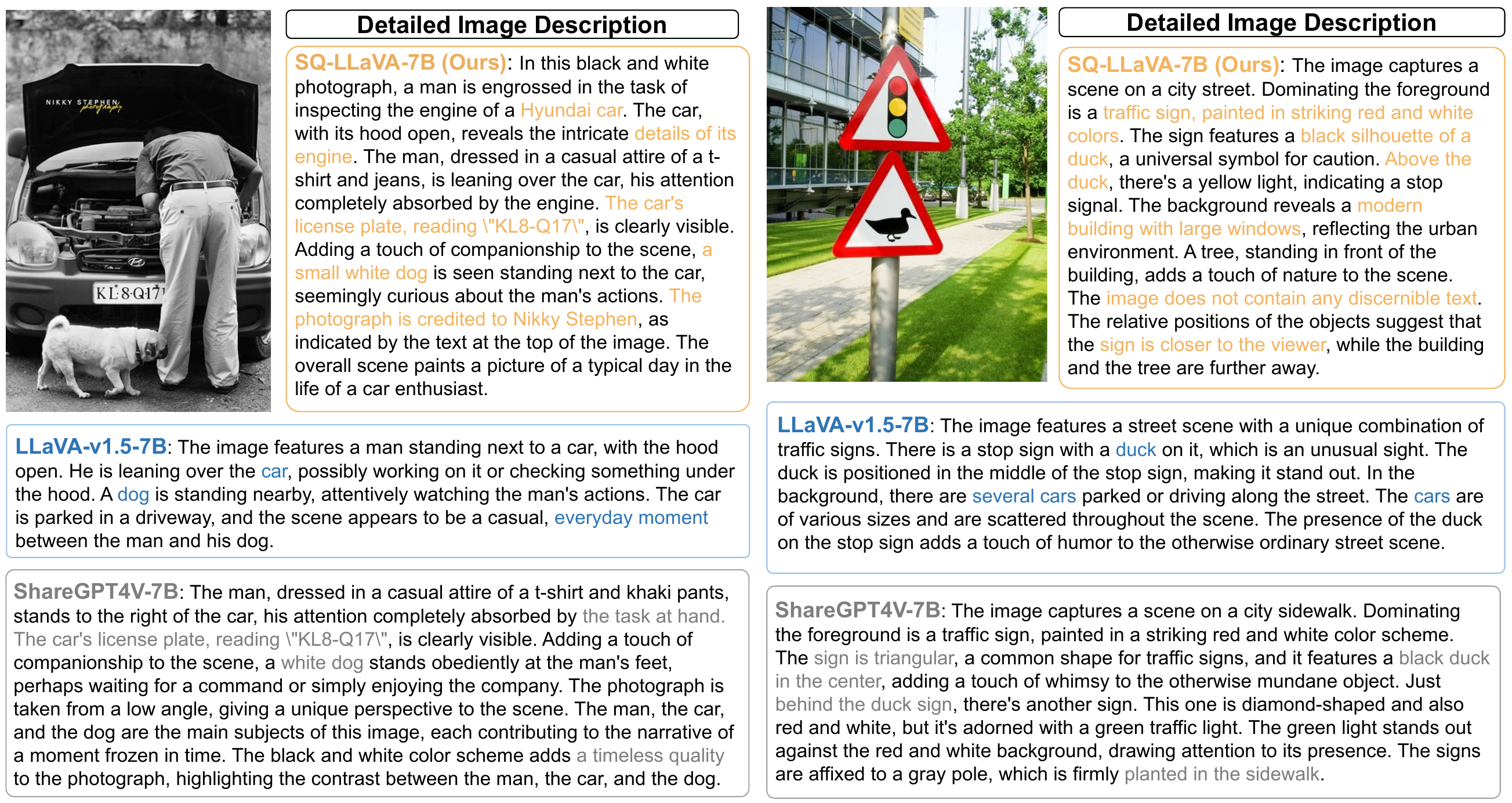}
  \caption{ \label{fig:captioning} A qualitative evaluation of detailed image descriptions from three models. We highlight the words and sentences that represent how each model describes the main object in the image. }
\end{figure}

The detailed image description example in Fig.~\ref{fig:captioning} demonstrates that SQ-LLaVA can generate visual information with concrete concepts within the image, such as ``Hyundai'', ``small white dog'', ``Nikky Stephen''. In contrast, LLaVA-v1.5 can only describe the image from a general concept, such as ``car'', ``dog'', or ``everyday moment''. Even though ShareGPT4V can generate detailed descriptions, it still suffers from the issue of object hallucination, such as ``perhaps waiting'' and ``behind the duck sign''. By observation, SQ-LLaVA can describe the image with less unintended text, yielding higher reliability. 

\noindent\textbf{Quantitative Analysis.}
SQ-LLaVA serves as a general-purpose vision-language model and enables zero-shot image captioning. As indicated in Table~\ref{tab:cap1}, SQ-LLaVA achieves $73\%$ and $66\%$ averaged improvement over ClipCap and DiscriTune on all datasets, indicating the effectiveness of visual instruction tuning. Compared with the baseline model LLaVA-V1.5~\cite{llava1-5}, SQ-LLaVA achieves $2\%$ averaged improvement on all datasets, indicating the effectiveness of visual self-questioning. Also, we find SQ-LLaVA$^*$ surpasses the baseline model ShareGPT4V~\cite{Chen2023ShareGPT4VIL} on Nocaps$^{out}$ and Conceptual dataset, demonstrating the adaptability of the proposed method on unseen testing images from new domains. Moreover, as shown in Table~\ref{tab:cap2}, SQ-LLaVA can be easily adapted to COCO captioning via instruction tuning on short descriptions.


\subsection{Ablation Study}

In Table~\ref{tab:ablation-vqa}, we conduct experiments with different architecture designs and training strategies on five question-answering benchmarks. For a fair comparison, we train the baseline models on our local machine with the same training recipe of LLaVA-LoRA~\cite{llava1-5}. 
Specifically, we present our full model and three ablated models by removing one component each time. We adopt the dataset~\cite{llava1-5} with $558$k for pre-training (PT) and $665$k for fine-tuning (FT). As compared, self-questioning (SQ) brings a consistent performance boost on all the five benchmarks, indicating the effectiveness of visual self-questioning on improving visual language understanding. Besides, we introduce the prototype extractor (Proto) to enhance visual representation, achieving $0.9\%$ improvement in average accuracy among five benchmark.
With all three components incorporated, we observe a $2.4\%$ improvement in average accuracy.

As shown by the bottom block of Table~\ref{tab:ablation-vqa}, we conduct experiments with the same ablation settings but with a larger scale of the visual instruction data~\cite{Chen2023ShareGPT4VIL} (\emph{i.e.}, for both PT and IT). Overall, SQ-LLaVA achieves $2.4\%$ improvement over the baseline model after training on the smaller dataset and achieves $3.0\%$ improvement after training on the larger dataset.

\begin{table*}[t]
\centering
\tabcolsep=0.3em
\caption{\label{tab:ablation-vqa} Ablation study of training strategy on visual instruction tasks. All models are in 7B scale with three components of ViT-LoRA (V-LoRA), self-questioning (SQ), and prototype extractor (Proto). We provide experiments on two instruction datasets~\cite{llava1-5,Chen2023ShareGPT4VIL} with different pre-training (PT) and instruction tuning (IT) data scales.   }
\scalebox{0.83}{
\begin{tabular}{llccc|ccccc|c}
\toprule
PT&IT & V-LoRA & SQ & Proto & VizWiz & SQA$^I$ & VQA$^T$ & POPE & LLaVA$^W$ &  Avg. \\ \midrule
\multirow{5}{*}{558K}&\multirow{5}{*}{665K} &  \xmark & \xmark &\xmark &49.4&68.4&58.2&86.5&67.1&65.9 \\
& &\cmark & \xmark & \cmark  & 52.4&67.9&58.6&87.7&65.6& 66.4\\
& &  \cmark & \cmark &\xmark  &  52.6&68.4&57.8&88.2&67.3&66.9 \\
& &  \xmark & \cmark &\cmark  &  53.4&69.3&58.1&87.9& 67.9&67.3 \\
& &  \cmark & \cmark &\cmark   & \cellcolor{cGrey}54.0 & \cellcolor{cGrey}68.9 & \cellcolor{cGrey}58.6 &\cellcolor{cGrey}87.7  & \cellcolor{cGrey}68.1 &\cellcolor{cGrey}67.5 \\ 
 \midrule
\multirow{5}{*}{1200K}&\multirow{5}{*}{700K} & \xmark & \xmark & \xmark  & 51.5&68.9&58.9&86.8&72.1&67.6  \\
& & \cmark & \xmark & \cmark  & 54.0&68.9&60.2&87.2&71.6 &68.4 \\
& & \cmark & \cmark & \xmark & 55.4&69.2&59.5&86.8&77.3&69.6 \\
& & \xmark & \cmark & \cmark &  54.2&70.3&60.5&87.5&72.7&69.0\\
& & \cmark & \cmark & \cmark &\cellcolor{cGrey}55.3  &\cellcolor{cGrey}70.5  &\cellcolor{cGrey}60.5  &\cellcolor{cGrey}87.2  &\cellcolor{cGrey}74.3  &\cellcolor{cGrey}69.6  \\ 
 \bottomrule
\end{tabular}}
\end{table*}


\section{Conclusions}
This work has introduced SQ-LLaVA, a new visual instruction tuning method that enhances general-purpose vision-language understanding and image-oriented question answering through visual self-questioning. Our experiments demonstrate that SQ-LLaVA achieves superior performance with fewer training parameters and instructional data. We have also conducted a comprehensive study on visual discovery/reasoning tasks and found that SQ-LLaVA generalizes well to a wide range of unseen tasks and outperforms several state-of-the-art methods. Qualitative assessments show that SQ-LLaVA strengthens visual representation and domain alignment, effectively reducing object hallucination and improving the semantic interpretation of images. Our findings highlight the potential of visual self-questioning as a powerful training strategy for the visual instruction tuning framework, paving the way for realizing more efficient and effective large vision-language models. Particularly, SQ-LLaVA frames questioning as an intrinsic goal of tuning LLMs, encouraging the exploration of the model's curiosity (the ability to ask questions proactively) in solving complex problems. 



%
%
\clearpage
\bibliographystyle{splncs04}
\bibliography{01393}
\end{document}